\documentclass[11pt,a4paper]{article}
\usepackage{authblk}
\usepackage[hyperref]{emnlp2020}
\usepackage{times}
\usepackage{latexsym}

\usepackage{balance}

\usepackage{microtype}
\usepackage{todonotes}
\usepackage{algorithm}
\usepackage{algorithmic}
\usepackage{amsmath}
\usepackage{subcaption}
\usepackage{multirow}
\usepackage[flushleft]{threeparttable}
\usepackage{caption}
\usepackage{subcaption}

\aclfinalcopy 

\title{MeDAL: Medical Abbreviation Disambiguation Dataset for Natural Language Understanding Pretraining}

\author[1]{\textbf{Zhi Wen}}
\author[1]{\textbf{Xing Han Lu}}
\author[1,2,3]{\textbf{Siva Reddy}}
\affil[1]{McGill University}
\affil[2]{Facebook CIFAR AI Chair}
\affil[3]{Mila -- Quebec Artificial Intelligence Institute}
\affil[ ]{\texttt {\{zhi.wen,xing.han.lu\}@mail.mcgill.ca}}
\affil[ ]{\texttt {siva@cs.mcgill.ca}}

\date{}

\begin{document}
\maketitle
\begin{abstract}
One of the biggest challenges that prohibit the use of many current NLP methods in clinical settings is the availability of public datasets. In this work, we present MeDAL, a large medical text dataset curated for abbreviation disambiguation, designed for natural language understanding pre-training in the medical domain. We pre-trained several models of common architectures on this dataset and empirically showed that such pre-training leads to improved performance and convergence speed when fine-tuning on downstream medical tasks.

\end{abstract}

\section{Introduction}

Recent work in mining medical texts focus on building deep learning models for different medical tasks, such as mortality prediction \citep{Grnarova2016NeuralPrediction} and diagnosis prediction \citep{Li2020InferringRecords}. However, because of the private nature of medical records, there are few large-scale, publicly available medical text datasets that are suitable for pre-training models, and real-world, private datasets are often small-scale and imbalanced. As a result, one of the biggest challenge in building deep learning-based NLP systems for biomedical corpora is the availability of public datasets \cite{Wang2018ClinicalReview}.

\begin{figure}[t]
    \centering
    \hspace{-1.5em}\includegraphics[width=0.51\textwidth]{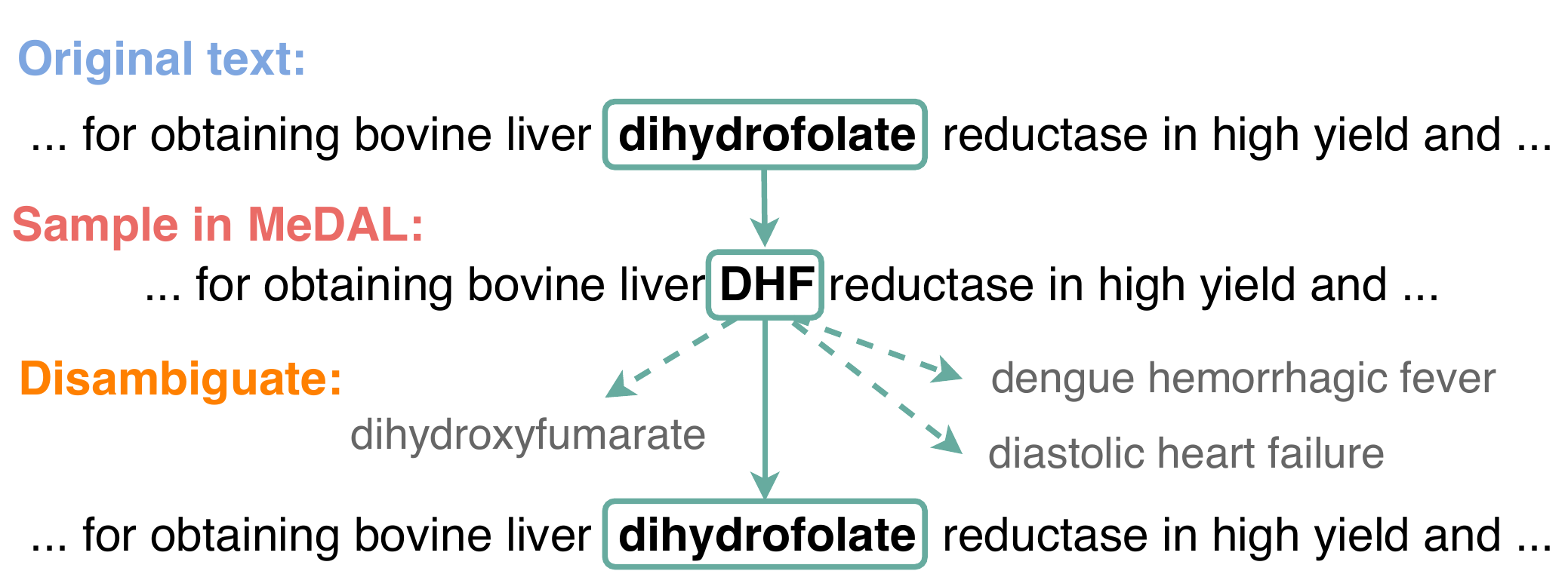}
    \caption{A sample in the MeDAL dataset.}
    \label{fig:rs-illustrate}
\end{figure}

\begin{figure*}[ht]
\centering
\includegraphics[width=0.8\textwidth]{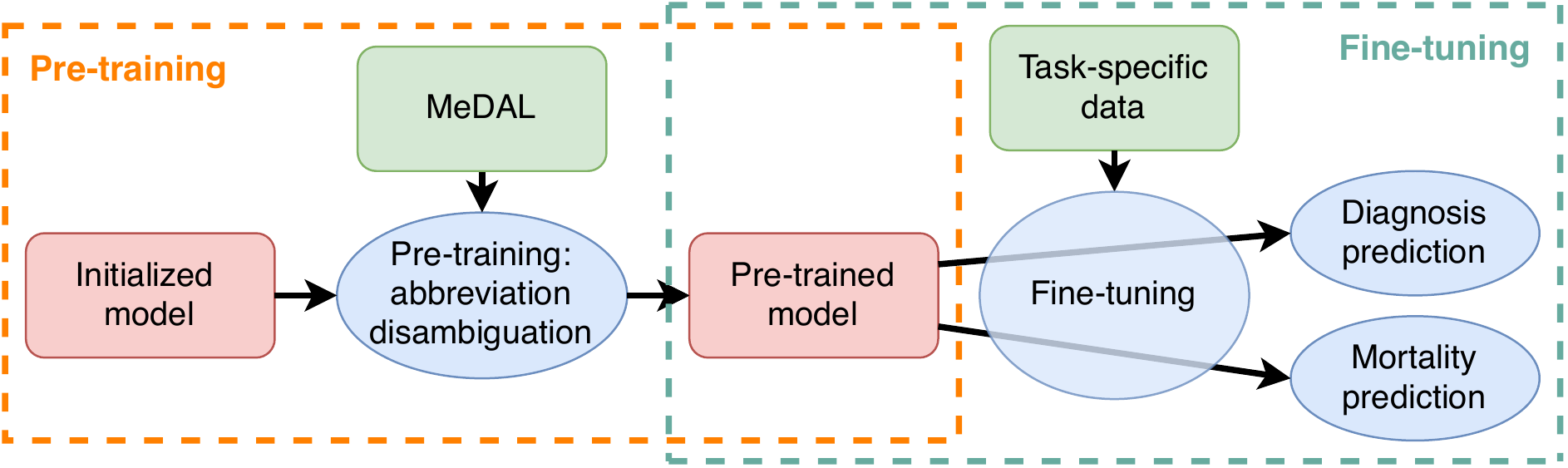}
\caption{\label{fig:diagram} Diagram of using MeDAL for pre-training NLU models in medical domain.}
\end{figure*}

To tackle this problem, we present \textbf{Me}dical \textbf{D}ataset for \textbf{A}bbreviation Disambiguation for Natural \textbf{L}anguage Understanding (MeDAL)\footnote{\href{https://github.com/BruceWen120/medal}{https://github.com/BruceWen120/medal}}, a large dataset of medical texts curated for the task of medical abbreviation disambiguation, which can be used for pre-training natural language understanding models. Figure \ref{fig:rs-illustrate} shows an example of sample in the dataset, where the true meaning of the abbreviation `DHF' is inferred from its context, and Figure \ref{fig:diagram} shows the pretraining framework. Although this dataset can be used for building abbreviation-expansion systems, its main purpose is to enable effective pre-training and improve performance on downstream tasks during fine-tuning.

The motivation behind using abbreviation disambiguation as the pre-training task is two-fold. First, abbreviations are widely used in medical records by healthcare professionals and can often be ambiguous \citep{Xu2007ANotes.,IslamajDogan2009UnderstandingAnalysis}.\footnote{For example, `MR' is a commonly used abbreviation which has a number of possible meanings, including `morphinone reductase', `magnetoresistance' and `menstrual regulation', depending on the context.} The ubiquitousness of abbreviations poses a restriction on building deep learning models for medical tasks, such as mortality prediction \citep{Grnarova2016NeuralPrediction} and diagnosis prediction \citep{Li2020InferringRecords}.

Second, we believe that understanding natural language in a knowledge-rich domain such as medicine requires understanding of domain knowledge at some level, similar to how humans can understand medical text only after receiving medical training. The abbreviation disambiguation task enables models to use domain knowledge to understand the global and local context, as well as the possible meanings of the abbreviation in the medical domain. 

Medical abbreviation disambiguation has long been studied \citep{Skreta2019TrainingDisambiguation, Li2019ADisambiguation, Finley2016TowardsData, Liu2018ExploitingExpansion,Joopudi2018AText, Jin2019DeepExpansion} and our work builds upon many of them. In particular, our data generation process is inspired by the reverse substitution technique \citep{Skreta2019TrainingDisambiguation, Finley2016TowardsData}. 

Our work differs from them in mainly two aspects. First, instead of trying to improve performance on abbreviation disambiguation itself, we propose to use it as a pre-training task for transfer learning on other clinical tasks. Second, existing datasets for medical abbreviation disambiguation, for instance CASI \citep{Moon2014AResources}, are small compared to datasets used for general language model pre-training, and as noted by \citet{Li2019ADisambiguation} some are erroneous. Thus, we chose to construct a new dataset large enough for effective pre-training.

Our main contributions are: a) we present a large dataset for pre-training on the task of medical abbreviation disambiguation. b) we provide empirical evidence of the benefit of abbreviation pre-training for a wide range of deep learning architectures.

\section{Abbreviation Disambiguation}

\subsection{Dataset Summary}\label{ss:dataset}


The MeDAL dataset consists of 14,393,619 articles and on average 3 abbreviations per article. The statistics of MeDAL are summarized in Table \ref{tab:abb-dataset-stats}.

\begin{figure}
    \centering
    \begin{subfigure}[h]{0.4\textwidth}
        \centering
        \includegraphics[width=\textwidth]{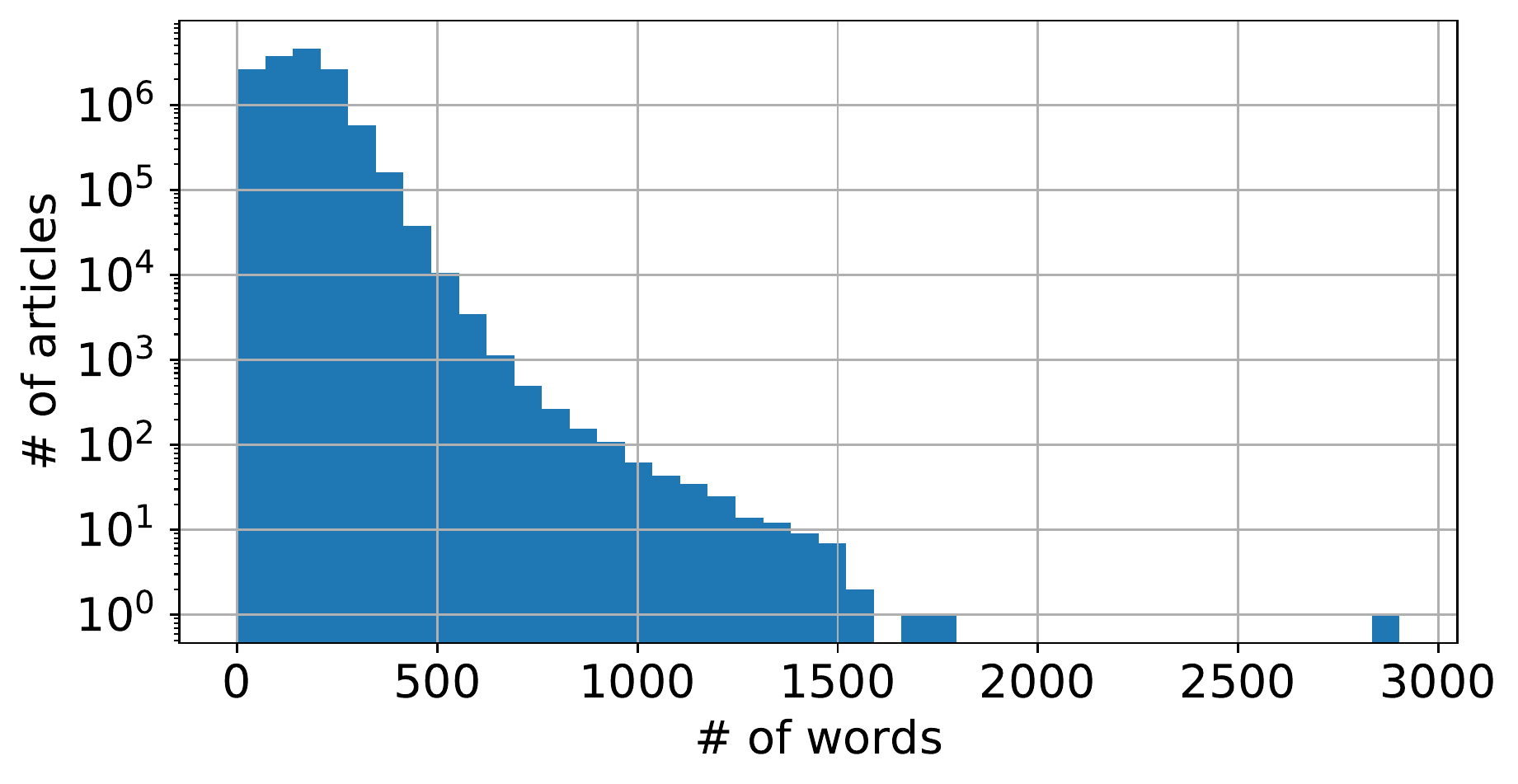}
        \caption{Word count distribution}
        \label{sfig:word_hist}
    \end{subfigure}
    \hfill
    \begin{subfigure}[h]{0.4\textwidth}
        \centering
        \includegraphics[width=\textwidth]{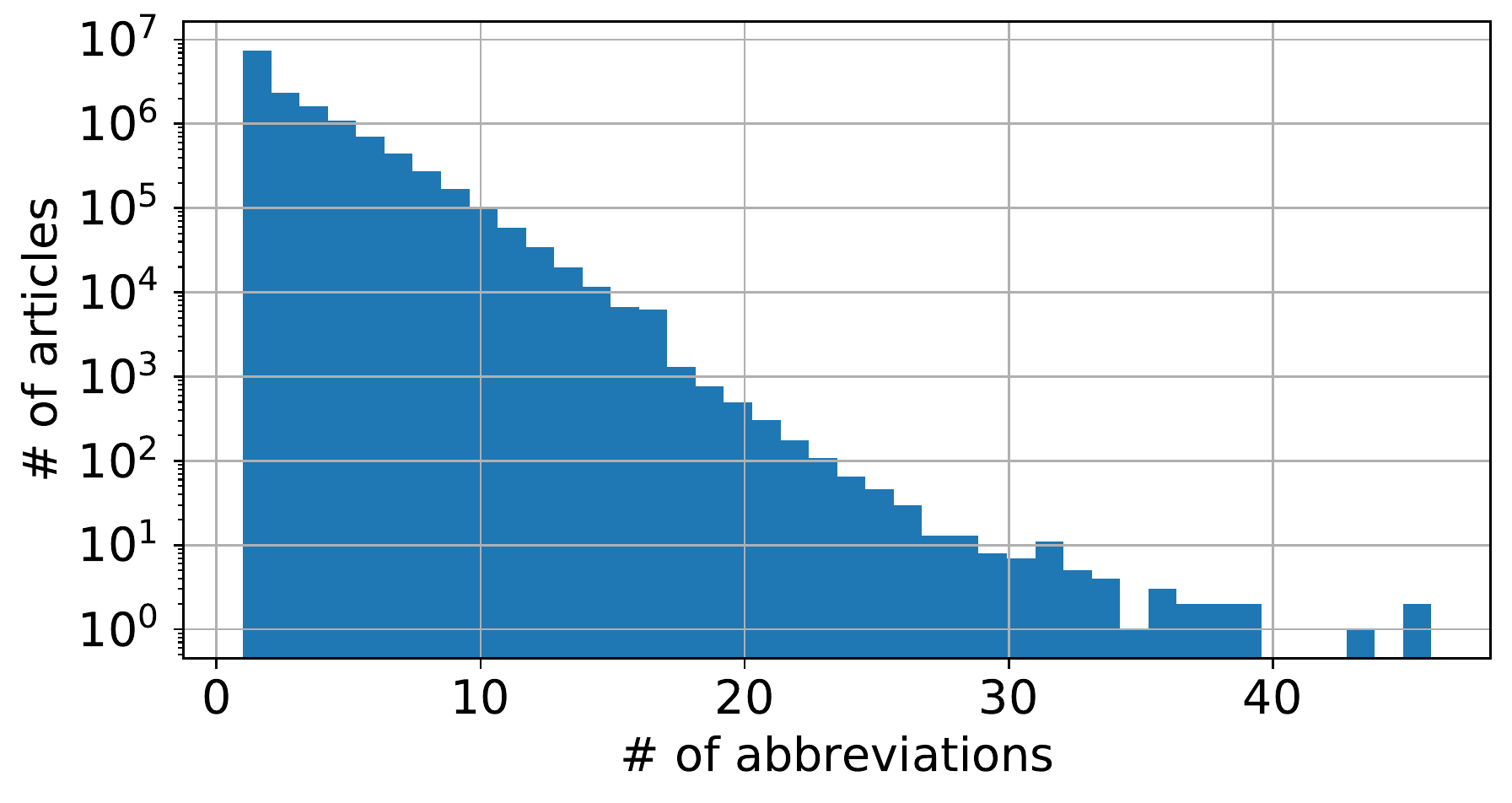}
        \caption{Abbreviation count distribution}
        \label{sfig:abb_hist}
    \end{subfigure}
    \caption{Distributions of number of words and number of abbreviations.}
    \label{fig:data_dists}
\end{figure}

The distribution of number of words and the distribution of number of abbreviations are shown in Figure \ref{sfig:word_hist} and Figure \ref{sfig:abb_hist}, respectively.

\subsection{Dataset Creation}

The MeDAL dataset is created from PubMed abstracts which are released in the 2019 annual baseline.\footnote{\href{https://www.nlm.nih.gov/databases/download/pubmed_medline.html}{https://www.nlm.nih.gov/databases/download/\\ pubmed\_medline.html}} PubMed is a search engine that indexes scientific publications in biomedical domain. The PubMed corpus contains 18,374,626 valid abstracts with 80 words in each abstract on average. 

We use reverse substitution \cite{Skreta2019TrainingDisambiguation} to generate samples without human labeling. We identify full terms in text that have known abbreviations and replace them with their abbreviations. For reverse substitution, mappings of abbreviations to expansions established by \citet{Zhou2006ADAM:MEDLINE} are used. Mappings where the abbreviation maps to only one expansion or the expansion maps to multiple abbreviations are discarded, resulting in 24,005 valid pairs of mappings. Among the valid mappings are 5,886 abbreviations, which means each abbreviation maps to about 4 expansions on average. 

To avoid completely removing all expansions and making them unseen to models, the expansions are substituted with a pre-defined probability. For our study, expansions are substituted with a probability of $0.3$, although our processing scripts allow for other values for future use. 

\subsection{Pretraining}

The task of abbreviation disambiguation is treated as a classification problem, where the classes are all possible expansions. 

Considering the huge size of the dataset and the associated computational cost, a subset of 5 million data points are sampled from the complete corpus, which are split into 3 million training samples, 1 million validation samples and 1 million test samples. This subset is used throughout this study.

When creating this subset, because the distribution of true expansions is highly imbalanced, a sampling strategy is adopted which essentially removes classes in increasing order of frequency in an iterative manner. The sampling strategy works in the following way: from each class label, $N_C = min(F_C, T)$ samples that have this label are randomly selected, where $F_C$ is the frequency of that class in the unsampled dataset, and $T$ is a threshold that is computed using Algorithm \ref{alg:threshold} such that each class can have at most $T$ samples, and $\sum_{C}N_C$ is equal to the total number of samples $N$.

The strategy iteratively removes classes, and at every iteration decreases $N'$ (which corresponds to the number of remaining samples) and $L$ (which corresponds to the number of labels remaining). Then, the rate $r$ is calculated based on how many classes $L$ can fit in the remaining $N'$ if each remaining $L$ has exactly $r$ samples. In this way, it is ensured that the moment the current class frequency $f_C$ being iterated is greater than the desired rate $r$, the sampling stops.

\begin{algorithm}
\caption{Compute threshold $T$}
\begin{algorithmic}
\REQUIRE array of class frequency $f$, $N>0$
\STATE Sort $f$ in increasing order
\STATE $L \leftarrow length(f)$
\STATE $N' \leftarrow N$
\FOR{each $f_C \in f$}
\STATE $N' \leftarrow N' - f_C$
\STATE $L \leftarrow L - 1$
\STATE $r = round(N' / L)$
\IF{$f_C \geq r$}
\RETURN $r + 1$
\ENDIF
\ENDFOR
\end{algorithmic}
\label{alg:threshold}
\end{algorithm}

\section{Evaluation Tasks}\label{s:eval-task}

\begin{table}[tp]
\centering
\begin{tabular}{|c|r|}
\hline
total \# of articles & 14,393,619 \\ \hline
median \# of words & 150 \\ \hline
mean \# of words & 152.47 \\ \hline
median \# of abbreviations & 2 \\ \hline
mean \# of abbreviations & 3.04 \\ \hline
\end{tabular}
\caption{Statistics of the MeDAL dataset}
\label{tab:abb-dataset-stats}
\end{table}

\paragraph{Mortality Prediction}

As a downstream task to evaluate models' performance in clinical settings, mortality prediction aims at predicting the mortality of a patient at the end of a hospital admission, using ICU patient notes. The mortality prediction dataset is generated from MIMIC-III \citep{Johnson2016MIMIC-IIIDatabase}. Medical notes in this MIMIC-III comprise of free-form text documents written by nurses, doctors, and many types of specialists, and are written throughout the patient's stay. Only notes written by physicians and nurses at least twenty-four hours before the end of the discharge time are used, for the goal is to accurately predict whether a patient is at risk of dying by the end of the admission. In order to balance positive and negative samples (roughly 10\% of patients expire at the end of an admission) while keeping as much text diversity as possible, we sample at most four notes from each surviving patient. 

The dataset generated has a total of 137,607 negative samples and 138,864 positively-labelled notes. Then, using stratified random splitting, we selected 75\%/10\%/15\% of the patients to be included in the training/validation/test splits. As an example of the ubiquitousness of abbreviations, `MR' appears 1,612 times in 1,366 samples in the test set alone. 

\paragraph{Diagnosis Prediction}

Similar to mortality prediction, diagnosis prediction aims to predict the diagnoses associated with a hospital admission from medical notes written during the admission. The same MIMIC-III medical notes and the same splits from mortality prediction are used, with seven training samples that have no diagnosis recorded removed. In MIMIC-III, diagnoses are recorded with International Classification of Diseases (ICD) codes, which are standardized codes designed for billing purposes. We discard minor distinctions of ICD codes under the same category by taking the first three digits (for codes that start with `E' or `V' the first four digits) of ICD codes.\footnote{For example, codes 4800 to 4809 represent viral pneumonia of different causes, and they are grouped into one ICD code 480.} After grouping, there are 1,204 unique diagnosis codes. 

Top-k recall is used for evaluation of models based on the similarities to real-life medical decision making \citep{Choi2015}, which is defined as the number of diagnosis codes in that admission that are present in the top k predictions of the model, divided by the number of diagnosis codes in that admission in total. Note that since most admissions have multiple diagnoses, a small k would result in a top-k recall less than 100\% even if all of the top k predictions are correct.\footnote{For instance, if an admission has 10 diagnoses codes, the highest possible top-5 recall for it would be $5/10=50\%$ which is when all of the top 5 predictions are correct.} On our dataset, the highest possible top-5, top-10 and top-30 recalls are $50.17\%$, $79.48\%$ and $99.88\%$ on validation set, and $49.75\%$, $79.23\%$ and $99.79\%$ on test set.

\begin{figure}[t]
    \centering
    \includegraphics[width=0.35\textwidth]{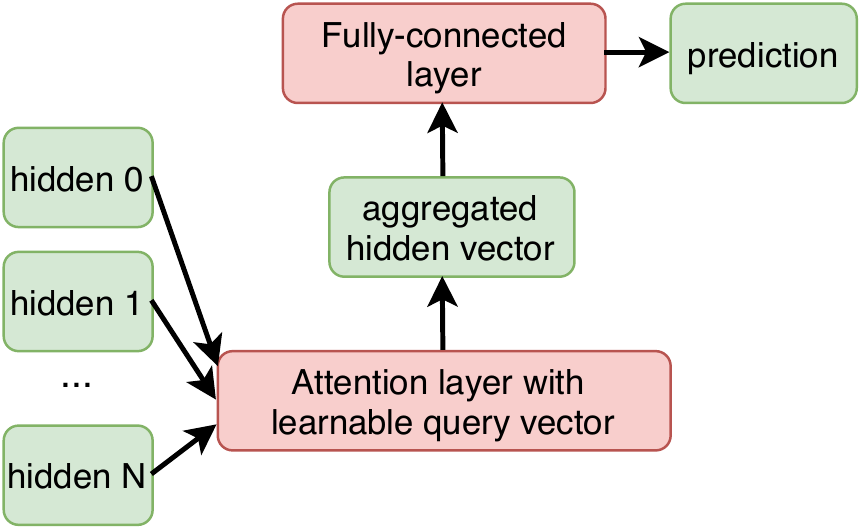}
    \caption{Attention output layer for mortality and diagnosis prediction.}
    \label{fig:mimic-out}
\end{figure}

\section{Models}\label{s:models}

The models are first pre-trained on the MeDAL dataset, then pre-trained weights are used to initialize models for training on the downstream tasks. We compared this training strategy with training respective models from scratch to validate the benefit of pre-training.

\paragraph{LSTM}

BiLSTM is used as a baseline model. Specifically, the BiLSTM consists of three layers with hidden size of 512. Pre-trained Fasttext model is used for word embeddings \citep{Bojanowski2017EnrichingInformation}.

\paragraph{LSTM + Self Attention}

To allow for leveraging information extracted by LSTM in a flexible manner, soft attention layers are added on top of LSTM. The attention layer is largely based on the soft attention by \citet{Bahdanau2014}. Its detailed formulation is included in Appendix \ref{app:attention}.

\paragraph{Transformers}

We used the pre-trained ELECTRA-small discriminator \citep{Clark2020ELECTRA:Generators} as an example of Transformer-based \citep{Vaswani2017} model and, since it was not pre-trained on medical text, we compared its performance with or without pre-training on abbreviation disambiguation. 

\paragraph{Task-specific Output Layer}

Depending on the task, the output layer can take various forms. For abbreviation disambiguation, the output layer is a fully-connected layer, whose input is the hidden vector at the location of the abbreviation from the previous layers and output space is all possible expansions. For mortality or diagnosis prediction which are not associated with any specific token, hidden vectors from the previous layers need to be first aggregated into one vector. This can be achieved by either a pooling layer or an additional attention layer with a learnable query vector. Then the output layer is a fully connected layer that takes the aggregated vector as input. The attention output layer is illustrated in Figure \ref{fig:mimic-out}. In preliminary experiments we found attention output layer generally improves models' performance compared to max-pooling output layer, and therefore it is used throughout the rest of the study unless otherwise noted.

\begin{figure}[t]
    \centering
    \includegraphics[width=0.45\textwidth]{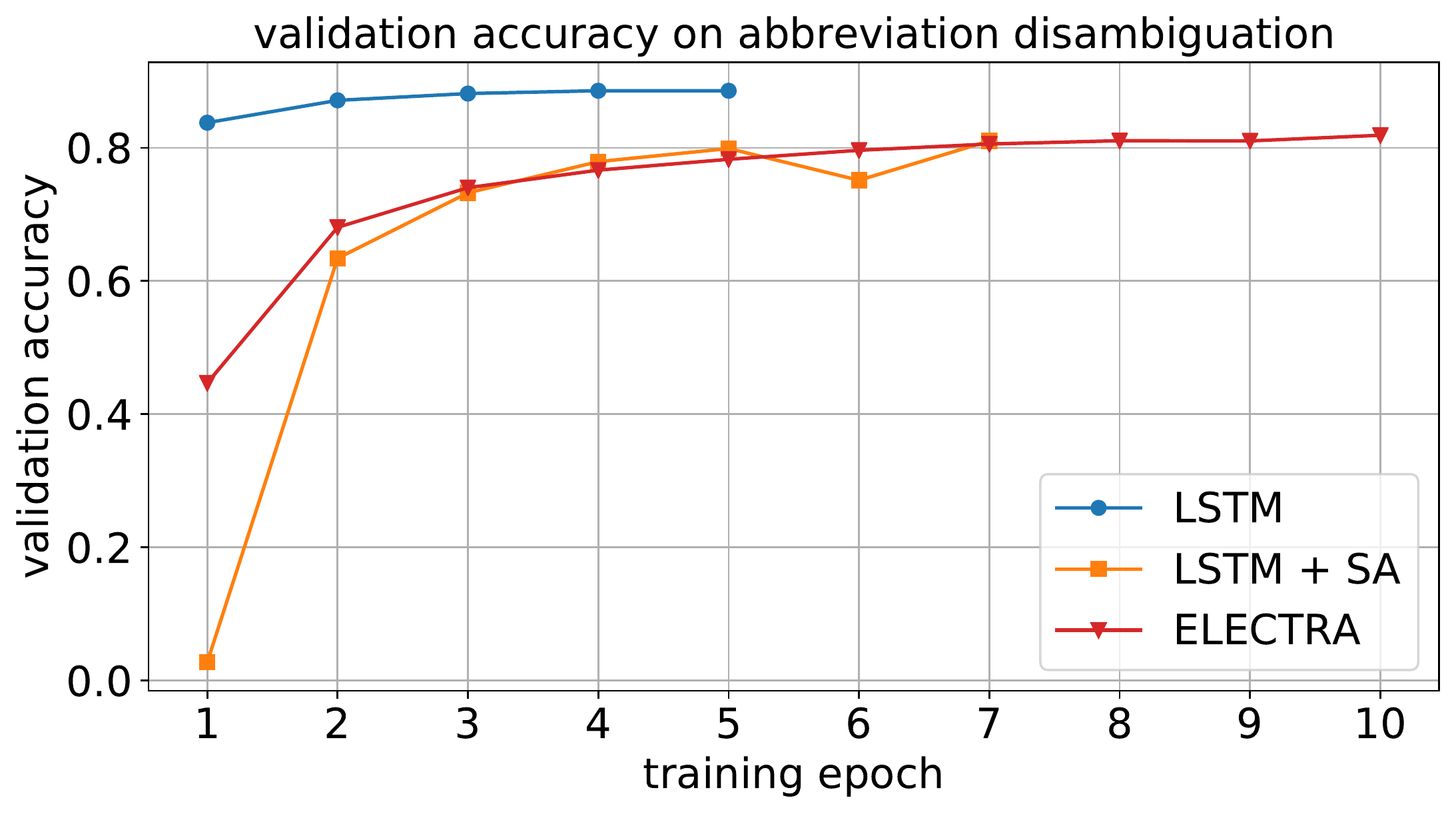}
    \caption{Validation accuracy on abbreviation disambiguation. `SA' stands for self attention layer.}
    \label{fig:abb-acc}
\end{figure}

\begin{figure}[tp]
    \centering
    \includegraphics[width=0.5\textwidth]{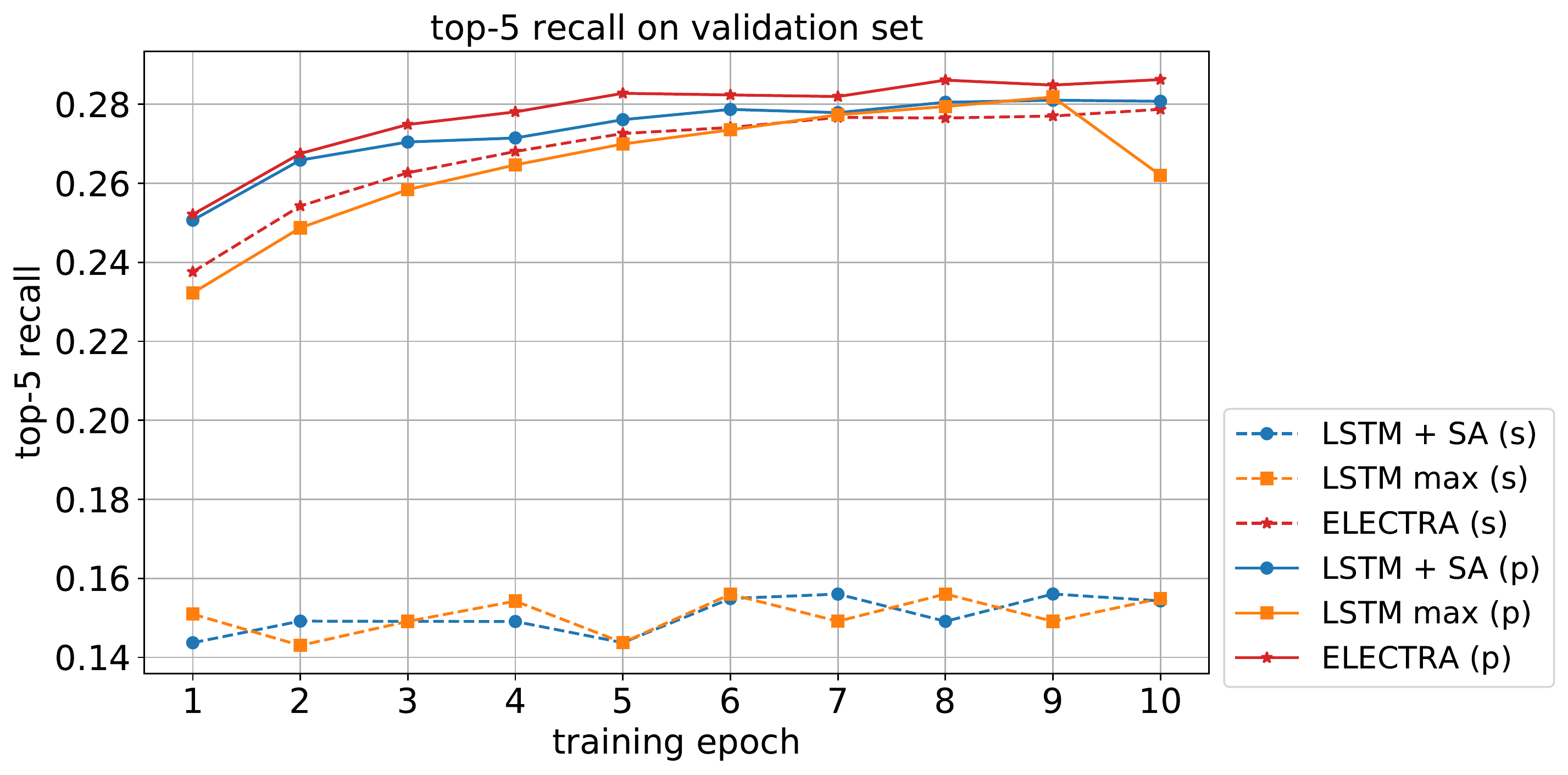}
    \caption{Top-5 recall on diagnosis prediction validation set. `SA' stands for self attention layer. `max' represents max-pooling output layer. `(s)' and `(p)' indicates whether the model is trained from scratch or pre-trained, respectively.}
    \label{fig:top-5-line}
\end{figure}

\begin{table}[t]
\centering
\begin{threeparttable}
\begin{tabular}{|c|c|c|}
\hline
\multirow{2}{*}{\textbf{Model}} & \multicolumn{2}{c|}{\textbf{Validation accuracy}} \\ \cline{2-3} 
 & \textbf{Pretrained} & \textbf{From scratch} \\ \hline
LSTM & \textbf{82.67\%} & 82.17\% \\ \hline
LSTM+SA & \textbf{82.46\%} & 80.29\% \\ \hline
ELECTRA & \textbf{84.19\%} & 83.92\% \\ \hline
 & \multicolumn{2}{c|}{\textbf{Test accuracy}} \\ \hline
LSTM & \textbf{82.80\%} & 82.61\% \\ \hline
LSTM+SA & \textbf{82.98\%} & 79.96\% \\ \hline
ELECTRA & \textbf{84.43\%} & 83.25\% \\ \hline
\end{tabular}
\caption{Results on mortality prediction. Bold font indicates the training strategy (pre-trained or from scratch) that has higher accuracy.}
\label{tab:mor-acc}
\end{threeparttable}
\end{table}

\section{Results}\label{s:exps}

Models' performance on the pre-training task, abbreviation disambiguation, is shown in Figure \ref{fig:abb-acc}. As the goal is not to optimize performance on this task, Figure \ref{fig:abb-acc} serves to confirm the models are properly pre-trained.

After pre-training, models are fine-tuned on the two downstream tasks to evaluate the benefit of pre-training. On the mortality prediction task, all three models that are pre-trained perform better than their from-scratch counterparts, shown in Table \ref{tab:mor-acc}.

The benefit of pre-training is more significant on diagnosis prediction, shown in Figure \ref{fig:top-5-line}. Both LSTM and LSTM + self attention perform considerably better if they pre-trained. In fact, the two models' performance increase by more than 70\% relatively. While for ELECTRA the gain is not as significant, pre-training leads to faster convergence during fine-tuning.

On the two downstream tasks, experiment results show that pre-training improves ELECTRA's performance even when the model is already fully pre-trained on non-medical texts and is among the state-of-the-art, and bring the other models' performance close to ELECTRA's. This shows that pre-training on the MeDAL dataset can generally improves models capabilities of understanding language in medical domain. The complete results can be found in Appendix \ref{app:add-results}.

\section{Conclusion and Discussion}

In this work, we present MeDAL, a large dataset on abbreviation disambiguation, designed for pre-training natural language understanding models in the medical domain. We pre-trained a variety of models using common architectures and empirically showed that such pre-training leads to improvement in performance as well as faster convergence when fine-tuning on two downstream clinical tasks.

\bibliographystyle{acl_natbib}
\bibliography{references}

\newpage

\appendix

\cleardoublepage
\balance

\section{Attention Layer}\label{app:attention}

Following \citet{Vaswani2017}, the attention layer can be expressed in terms of key, query and value vectors, denoted as $\mathbf{k}_i$, $\mathbf{q}_i$ and $\mathbf{v}_i$ respectively, where the subscript $i$ denotes the location in the sequence. Specifically, the attention layer in our models is defined as Equation \ref{eq:weight}.

\begin{equation}\label{eq:weight}
    w_{ij}=\frac{\exp \alpha_{ij}}{\sum_n\exp \alpha_{in}}
\end{equation}

$\alpha_{ij}$ in Equation \ref{eq:weight} is computed with Equation \ref{eq:alpha}, where $W_a$ and $b$ are learnable parameters.

\begin{equation}\label{eq:alpha}
    \alpha_{ij}=\tanh(\mathbf{q}_i\cdot W_a \cdot {\mathbf{k}_j}^T+b)
\end{equation}

Here $w_{ij}$ is the weight assigned to location $j$ for location $i$. Then the output of the attention layer at location $i$ is computed by taking the weighted sum of value vectors at all locations, i.e. $\mathbf{o}_i=\sum_n w_{in} \cdot \mathbf{v}_n$, where $\mathbf{o}_i$ denotes the output of attention layer at location $i$. Unless otherwise noted, throughout this paper $\mathbf{k}_i$, $\mathbf{q}_i$ and $\mathbf{v}_i$ are all equal to the hidden vector at position $i$ from the previous layer $\mathbf{h}_i$.

\section{Experiment Details}\label{app:exp-setup}

Except for ELECTRA, the rest of the models are trained with Adam optimizer \citep{Kingma2014} with learning rate of $0.001$. Text is tokenized using pre-trained Fasttext embeddings \citep{Bojanowski2017EnrichingInformation}. All LSTM modules are bi-directional and have 3 layers, with hidden size of 512. Batch size is set to 64. We experimented with various choices of batch sizes, including 32, 64, 96 and 128, and noted only minimal differences. ELECTRA is trained with Adam optimizer with learning rate of $0.00002$ and with batch size of 16.

\section{Additional Experiments Results}\label{app:add-results}

\begin{figure}[hp]
    \centering
    \includegraphics[width=0.5\textwidth]{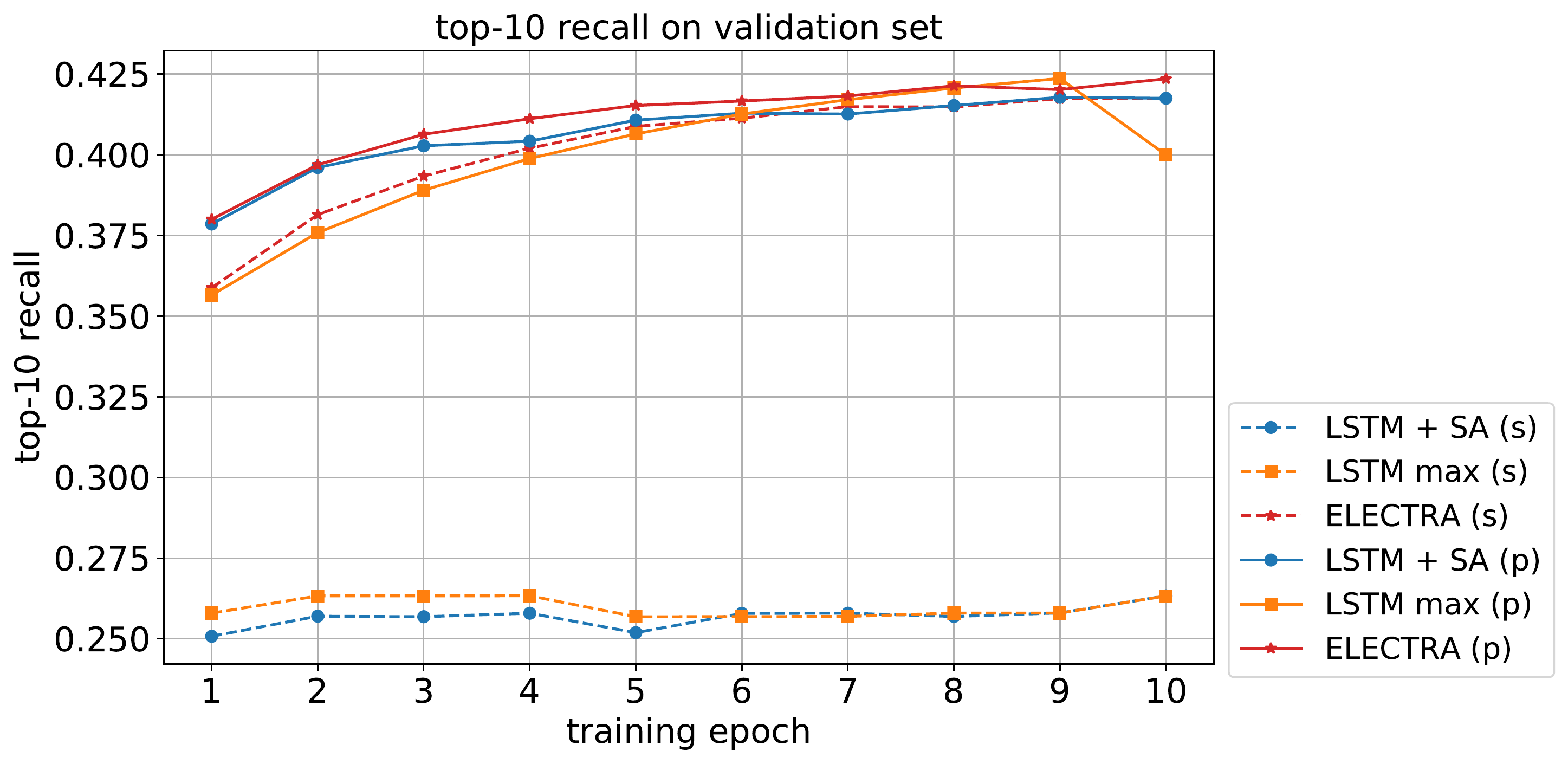}
    \caption{Top-10 recall on diagnosis prediction validation set. `SA' stands for self attention layer. `max' represents max-pooling output layer. `(s)' and `(p)' indicates whether the model is trained from scratch or pre-trained, respectively.}
    \label{fig:top-10-line}
\end{figure}

\begin{figure}[hp]
    \centering
    \includegraphics[width=0.5\textwidth]{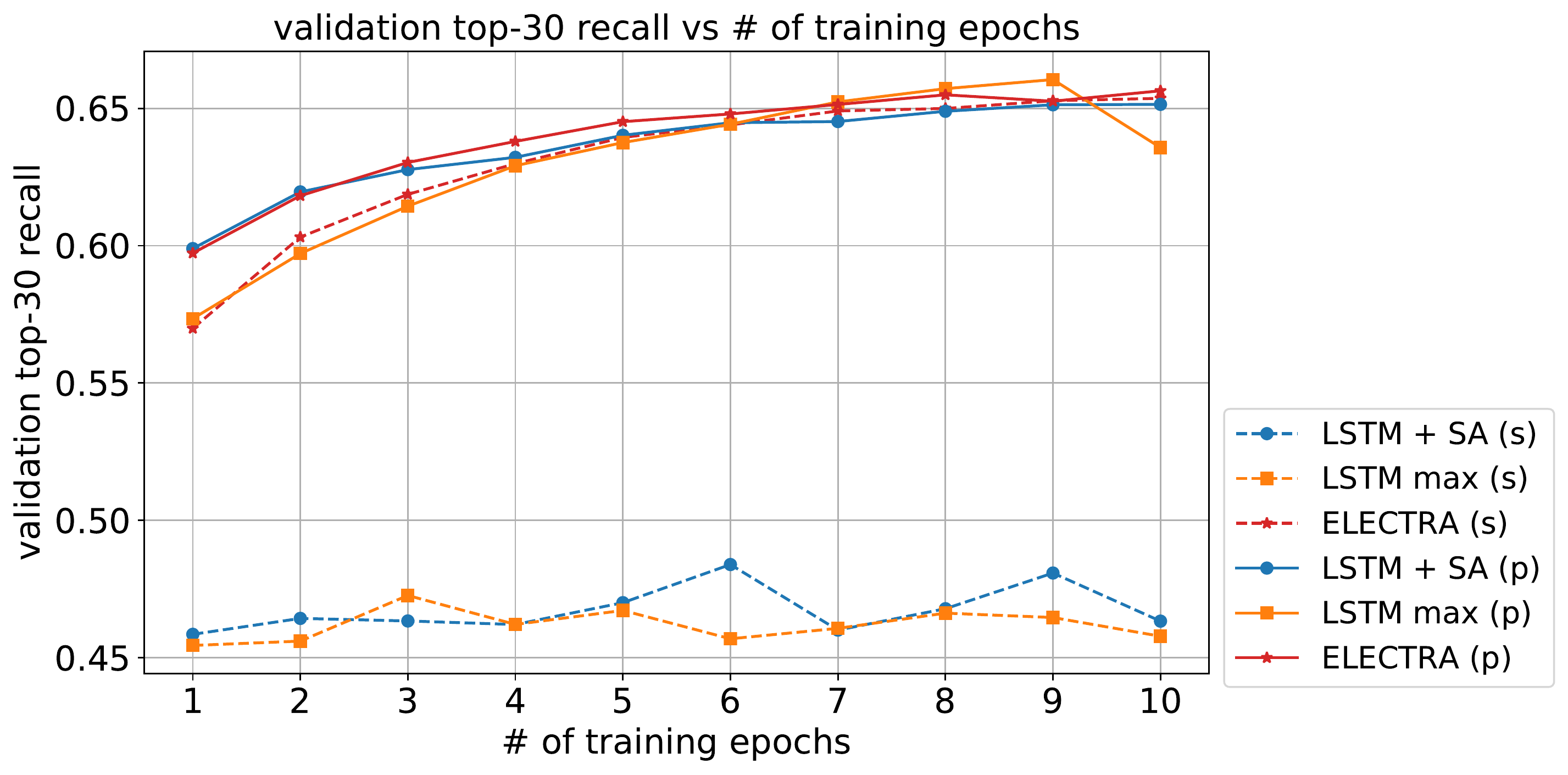}
    \caption{Top-30 recall on diagnosis prediction validation set. `SA' stands for self attention layer. `max' represents max-pooling output layer. `(s)' and `(p)' indicates whether the model is trained from scratch or pre-trained, respectively.}
    \label{fig:top-30-line}
\end{figure}

Figure \ref{fig:top-10-line} to Figure \ref{fig:top-30-line} show the top-10, and top-30 recalls on diagnosis prediction, respectively. Table \ref{tab:diag-recalls} shows the complete performance of models on diagnosis prediction.

\begin{table*}[ht]
\centering
\begin{threeparttable}
\begin{tabular}{|c|c|c|c|c|c|c|}
\hline
\multirow{2}{*}{\textbf{Model}} & \multicolumn{6}{c|}{\textbf{Validation performance}} \\ \cline{2-7} 
 & \multicolumn{2}{c|}{\textbf{Top-5 recall}} & \multicolumn{2}{c|}{\textbf{Top-10 recall}} & \multicolumn{2}{c|}{\textbf{Top-30 recall}} \\ \hline
 & Pre-trained & From scratch & Pre-trained & From scratch & Pre-trained & From scratch \\ \hline
LSTM & \textbf{26.20\%} & 15.49\% & \textbf{40.00\%} & 26.33\% & \textbf{63.57\%} & 45.78\% \\ \hline
LSTM+SA & \textbf{28.08\%} & 15.43\% & \textbf{41.75\%} & 26.33\% & \textbf{65.15\%} & 46.33\% \\ \hline
Electra & \textbf{28.63\%} & 28.08\% & \textbf{42.35\%} & 41.74\% & \textbf{65.64\%} & 65.37\% \\ \hline
 & \multicolumn{6}{c|}{\textbf{Test performance}} \\ \hline
LSTM & \textbf{26.94\%} & 15.67\% & \textbf{40.59\%} & 25.97\% & \textbf{65.49\%} & 45.15\% \\ \hline
LSTM+SA & \textbf{27.47\%} & 15.93\% & \textbf{41.24\%} & 25.97\% & \textbf{65.86\%} & 45.67\% \\ \hline
Electra & 27.88\% & \textbf{27.90\%} & 41.76\% & \textbf{41.82\%} & 66.23\% & \textbf{66.49\%} \\ \hline
\end{tabular}
\caption{Performance on diagnosis prediction\tnote{a}\tnote{b}}
\label{tab:diag-recalls}
\begin{tablenotes}
\small
\item[a] Note that, as discussed in Section \ref{s:eval-task}, on our dataset the highest possible top-5, top-10 and top-30 recalls are $50.17\%$, $79.48\%$ and $99.88\%$ on validation set, and $49.75\%$, $79.23\%$ and $99.79\%$ on test set.
\item[b] Bold font indicates the training strategy (pre-trained or from scratch) that has higher accuracy.
\end{tablenotes}
\end{threeparttable}
\end{table*}

\end{document}